\documentclass[preprint]{elsarticle}

\usepackage{graphicx}
\usepackage{wasysym}
\usepackage{amsfonts}
\usepackage{amssymb}
\usepackage{xcolor}
\usepackage[T1]{fontenc}
\usepackage[a4paper, total={6in, 9in}]{geometry}

\renewenvironment{abstract}
 {\small
  \begin{center}
  \bfseries \abstractname\vspace{-.5em}\vspace{0pt}
  \end{center}
  \list{}{%
    \setlength{\leftmargin}{20mm}
    \setlength{\rightmargin}{\leftmargin}%
  }%
  \item\relax}
 {\endlist}
\newcommand{\comment}[1]{}

\begin{document}

\title{A Role of Environmental Complexity on Representation Learning in Deep Reinforcement Learning Agents}

\author{Andrew Liu}
\ead{aliu@math.utah.edu}
\author{Alla Borisyuk}
\ead{borisyuk@math.utah.edu; Dept. of Mathematics, 155 E 1400 S, Salt Lake City, UT 84109, USA}

\maketitle{}

\begin{abstract}
We developed a simulated environment to train deep reinforcement learning agents on a shortcut usage navigation task, motivated by the Dual Solutions Paradigm test used for human navigators. We manipulated the frequency with which agents were exposed to a shortcut and a navigation cue, to investigate how these factors influence shortcut usage development. We find that all agents rapidly achieve optimal performance in closed shortcut trials once initial learning starts. However, their navigation speed and shortcut usage when it is open happen faster in agents with higher shortcut exposure. 
Analysis of the agents' artificial neural networks activity revealed that frequent presentation of a cue initially resulted in better encoding of the cue in the activity of individual nodes, compared to agents who encountered the cue less often.
However, stronger cue representations were ultimately formed through the use of the cue in the context of navigation planning, rather than simply through exposure.
We found that in all agents, spatial representations develop early in training and subsequently stabilize before navigation strategies fully develop, suggesting that having spatially consistent activations is necessary for basic navigation, but insufficient for advanced strategies.
Further, using new analysis techniques, we found that the planned trajectory rather than the agent's immediate location is encoded in the agent's networks. Moreover, the encoding is represented at the population rather than the individual node level. These techniques could have broader applications in studying neural activity across populations of neurons or network nodes beyond individual activity patterns.

\end{abstract}


\section{Introduction}

Navigation is a critical and complex skill that is essential for human and animal success in the world.
It involves tracking one's position, understanding landmarks, recognizing routes, and acquiring survey knowledge \cite{siegel:1975} to plan a path towards a goal.
Landmarks are salient features of the environment that can be used to locate or orient, and routes are familiar paths between locations, often incorporating landmarks.
Survey knowledge, akin to a cognitive map, allows planning new routes and depends on integrating diverse information sources.
Success in developing survey knowledge varies considerably between individuals \cite{weisberg:2018}.

\begin{figure}
\centering
\includegraphics{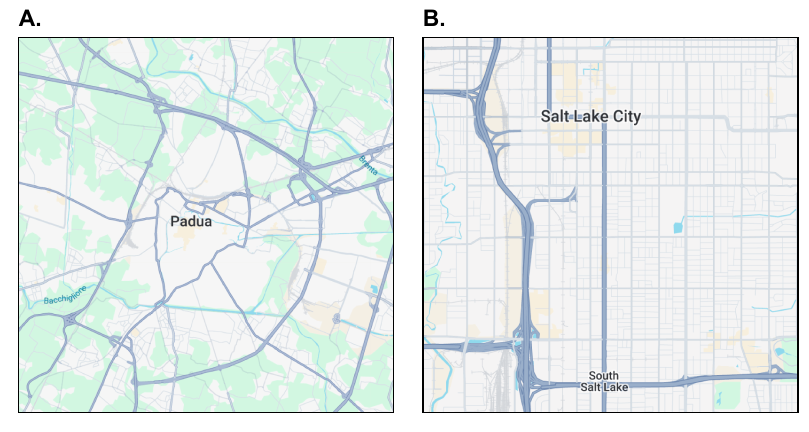}
\caption{
Maps of \textbf{A.} Padua (Veneto, Italy) and \textbf{B.} Salt Lake City (Utah, USA), taken from Google Maps \cite{googlemaps}
}
\label{fig:slc_padua}
\end{figure}

There are many factors that can contribute to individual differences in navigation skill, including, e.g., sex \cite{nazareth:2019} and age \cite{merhav:2019}.
Recently it has been suggested \cite{barhorst:2021} that individual's home environment also helps to shape their navigation strategies and skills. Barhorst-Cates et al. \cite{barhorst:2021} compared navigation performance between people from Salt Lake City (Utah, USA) and those from Padua (Veneto, Italy). These two cities provide contrasting environments (sample maps are shown in Fig \ref{fig:slc_padua}): Salt Lake City has grid-like streets and surrounding mountains provide global cues and Padua has organic, maze-like streets.
It was found that individuals from Padua generally had better access to survey knowledge, ability to use cues, and overall navigation skill. One of the tasks that was used in this study is the dual-solutions paradigm (DSP) \cite{marchette:2011}, which assesses how often shortcuts are used during navigation in a maze. In DSP the tendency to use shortcuts is thought to indicate the use of survey knowledge by the individual, and it has been used in both rodent and human studies.

Inspired by the human studies, we developed a toy version of these navigation tasks to study the effect of the ``home'' environment, the shortcut usage, and the attendant ``survey knowledge'' in the artificial agents (in Section \ref{sec:simulated_environment}). We applied deep reinforcement learning (RL) \cite{sutton:2018} to model the navigation learning process in artificial agents. The agents were trained with different availability
of a visual landmark and a shortcut, resembling different home environments for humans, 
and then were tested in a toy version of the dual-solution paradigm task. We examined their navigation strategies and shortcut use, and analyzed the resulting activities in their artificial neural networks for representations of survey knowledge and other components of navigation decisions. 

Our dual objectives are to leverage understanding of human navigation to construct an informative and constrained training environment for artificial agents, and to develop tools for probing and analyzing the network activities (representations) at different stages of the agents' training.

RL has been successfully applied to various complex navigation tasks, providing a robust challenge for testing agent capabilities \cite{kempka:2016, mirowski:2016, mnih:2016}.
In our approach \cite{liu:2024} we use simplified versions of the navigating environments, to enable a more nuanced understanding of the learning process, the resulting strategies, and the analysis of the neural network representations in navigating agents \cite{liu:2024}. 

Representations can be viewed as correlations between the state of the environment and the activity of the agent's neural network.
In other words, they act as an internal model of the world that the agent faces.
In this work, we first looked at the spatial representations recorded from individual nodes. There are similar to activities of place cells found in the hippocampal system \cite{moser:2015}.
We find that these encodings are foundational to any navigation strategy and develop early in training.
In contrast, landmark encoding, which is measured as activity in nodes that differentiate between the presence and absence of a given visual cue,  continues to develop throughout the training period.
In particular, the landmark encoding is stronger in agents who are given more access to landmarks early on, but continue to develop more robustly in agents that learn to apply them as part of their navigation strategy, independent of the frequency that agents encounter the cues.

Beyond individual neural network nodes, we also looked at two representations coded across populations of nodes.
First, we show that clustering the collection of spatial activity heatmaps from all network nodes can be used to reveal what salient features of the environment are being encoded by the network. Individual node activity patterns can be noisy and hard-to-interpret. Our clustering analysis allows to look beyond the noise to reveal the features of the environment that are available to the agent. This analysis can serve as a useful exploration tool to understand activity in brains or artificial neural networks of agents.
Second, we find that agents are able to effectively encode the trajectory that they are on, suggesting an intent or planned path, using a population representation.
This planned path representation is continuously refined throughout the training process. It is also independent of cue availability, which we can show by making use of our ability to manipulate the visual cues given to agents during navigation.

While this experimental set up is much abstracted from the more complex navigation world that real people and animals face, the results may still provide some insight into biological navigators.
Our observations on landmark sensitive nodes suggest suggest that individuals from Padua who may have less consistent landmarks than those in Salt Lake City nevertheless can develop stronger internal cue representations due to their increased survey knowledge of experiencing difficult navigating conditions.
From the variations of our toy model, we observed significant changes in learning efficiency in artificial agents triggered by relatively small changes to the environment.
We propose a modification to the DSP where a few demonstrations of shortcuts are given initially and hypothesize that this adjustment would strongly boost the frequency of shortcut usage in those subjects who might otherwise be route followers.

Overall, we find that even relatively simple navigation environments can give rise to interesting behavior and representation learning dynamics in artificial agents.
These trained neural networks can provide insight into the internal cognition of real navigators, and suggest interventions to affect skill and knowledge development.

\section{Methods}

\subsection{Reinforcement Learning}

We train deep RL \cite{sutton:2018} artificial agents to perform a navigation task, using a set up and training approach similar to ones we used in an earlier publication \cite{liu:2024}.
The agents are trained in an environment that can be conceptualized as a Partially Observable Markov Decision Process (POMDP) defined by a $(S, A, P, R, \Omega, O)$ tuple.
At each time step $t$, the environment has state $s_t \in S$ and gives a limited amount of information about that state to the agent in an observation $o_t \in \Omega$.
The mapping $O: S \rightarrow \Omega$ defines what information is given.
The agent performs actions from a discrete action space $a_t \in A$, the state evolves according to a transition function $P: S \times A \rightarrow S$, and a reward $r_t$ is given according to $R : S \times A \rightarrow \mathbb{R}$.

The agent learns a policy $\pi_\theta(a_t | o_t, h_t) = \mathbb{P}[a_t=a | o_t=o, h_t=h]$ where $h_t$ is a hidden state $h_t \in \mathbb{R}^k$ used in a recurrent layer of the network.
The policy is parameterized by a neural network with parameters $\theta$.
Simultaneously, the agent learns to predict expected return known as the value $V(o_t, h_t) = \mathbb{E}^{\pi_\theta}[G_t|o_t,h_t]$.
The return $G_t = \sum^\infty_{k=0} \gamma^k r_{t+k+1}$ is a discounted sum of future rewards, with discount factor $\gamma \in  [0, 1)$.

\subsection{The simulated navigation environment}
\label{sec:simulated_environment}

\begin{figure}
\centering
\includegraphics{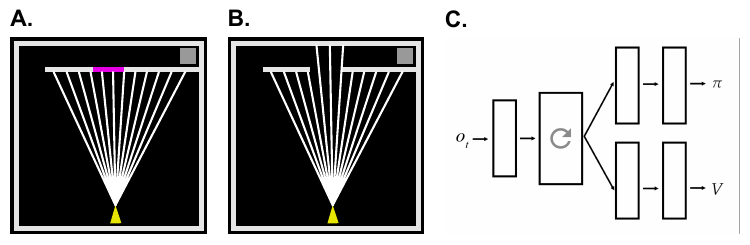}
\caption{
\textbf{A and B.} Depictions of the simulated shortcut navigation environment with \textbf{A.} shortcut closed and \textbf{B.} shortcut opened.
All walls are colored white except for the shortcut wall when closed, which is pink.
A gray box in the very top-right corner represents the navigation target.
The agent is depicted as a yellow triangle, with white lines extending outwards its sight lines.
\textbf{C.} A schematic of the neural network that agents are parameterized by.
Each block represents network layer of 64 nodes, and the arrows represent the activations of one layer being passed to the next in a fully-connected weighted sum.
The block with a circled arrow is a recurrent layer.
Observation are input ($o_t$ on the left of the schematic), and the network splits into actor ($\pi$) and critic ($V$) branches.
}
\label{fig:env_and_nn}
\end{figure}

In this study, we train agents to perform an episodic \cite{sutton:2018} navigation task (simulations run in Python), depicted in Fig \ref{fig:env_and_nn}A and \ref{fig:env_and_nn}B.
The agent's goal is to navigate to an invisible target located in the top-right corner of the maze, represented as a gray box in the drawing, at the end of a corridor at the top of the environment.
A key feature of this environment is the pink wall in the center of the corridor (Fig \ref{fig:env_and_nn}A).
This wall can be opened (Fig \ref{fig:env_and_nn}B), creating a shortcut in the maze and simultaneously removing the primary salient landmark in the environment.
At the start of the episode, the shortcut is randomly chosen to be open with probability $p$, where $p$ serves as the key parameter of the environment.
Varying this parameter gives rise to variations of the environment, and each agent is trained with a fixed $p$ throughout all episodes.

Sight lines (visualized in Fig \ref{fig:env_and_nn}) return both the color and distance to the wall each line intersects.
Color is encoded as a one-hot vector with two possibilities (white or pink).
This information forms a 36-dimensional observation vector $o_t \in \Omega = \mathbb{R}^{36}$, as each of the 12 vision lines provides one scalar value (distance) and two dimensions for color.
At each time step, the agent picks from four possible actions $a_t$: a left or right turn of $0.2$ radians, a forward movement of $10$ units, or no movement.
The navigation arena is a $[0, 300] \times [0, 300]$ square.

Each episode resets either when the agent reaches the goal or when $200$ time steps elapse.
If the goal is reached, a reward $r_t$ of $1$ is earned, and the reward is $0$ for all other time steps.
Upon resetting, the starting position of the agent is randomized such that the starting height is in $[0, 240]$ to ensure that it starts below the corridor, which is the area above the maze's long wall, 50 units high from the top edge.
An initial direction is also selected uniformly at random in $[0, 2\pi)$.
After initialization, the environment evolves deterministically according to actions selected by the agent.


\subsection{Agent network and training algorithm}

Fig \ref{fig:env_and_nn}C shows a schematic of the artificial neural network that the agent is trained with.
It is made up of primarily feed-forward layers, where the activation of one layer is propagated forward as a weighted sum to the next.
The network splits into actor and a critic branches that output $\pi$ and $V$ respectively.
The layer before the split is a gated recurrent unit \cite{cho:2014}, which is a recurrent layer that adds memory capabilities to the network, and is depicted with a circled arrow.
Each layer in the network consists of 64 nodes.
The agent accumulates training data by executing its policy in the environment and updates its neural network weights using a policy gradient method known as Proximal Policy Optimization (PPO) \cite{pytorchrl:2018, schulman:2017}.
Initially, each new agent starts with a randomly weighted neural network, leading to initially random actions.
The output from the actor branch is a four-dimensional vector, transformed into a probability distribution over the four potential actions using a softmax function.
The critic branch outputs a single scalar which is the agent's predicted future return from the current state.

\subsection{Learning curves and initial non-learning periods}
\label{sec:non_learning_period}

\begin{figure}
\centering
\includegraphics{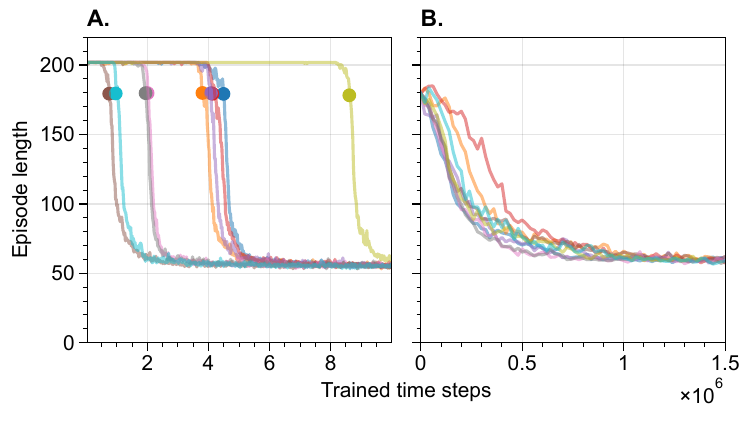}
\caption{
Examples of shifting learning curves, for agents trained with $p=0.1$. 
Each line represents one agent. 
\textbf{A.} Original learning curves showing mean episode length for an agent.
Dots depict where the mean first falls below 180.
\textbf{B.} Learning curves shifted such that they start at the dots from \textbf{A}.}
\label{fig:shift_learning_curves}
\end{figure}

Learning curves graphically represent the performance of agents based on their cumulative experienced time steps, shown in Fig \ref{fig:shift_learning_curves}A.
These show mean episode length for individual agents, where an episode length of 200 indicates that the time limit was reached and the agent failed to reach the target.
At the start of training, agents perform random actions due to their randomly initialized neural networks.
The first meaningful learning signals occur when agents first reach the goal, from which they quickly learn to replicate successful trajectories.
The learning curves show a steep improvement in performance as the agents develop effective strategies.

This pattern leads to an initial non-performant period which has random length, and its length does not impact learning dynamics afterwards.
It is easier for an agent taking random actions to reach the platform when the shortcut is open.
Hence, agents trained with a higher probability $p$ of the shortcut being open typically experience a shorter non-performant period.
We disregard this initial variability by looking for the first time an agent's average episode length drops below 180 time steps, shown by dots in Fig \ref{fig:shift_learning_curves}A. 
Fig \ref{fig:shift_learning_curves}B adjusts learning curves to start from this point, providing a clearer view of learning progression.
Each agent undergoes training for $6e6$ time steps beyond this initial point, and all results in this paper will use these shifted starting points.

\subsection{Evaluation episodes}
\label{sec:evaluation_episodes}

To track agent performance and learning efficiency in a consistent manner, we employ a suite of standardized evaluation trials.
First, we randomly generate 50 initial positions and directions, which are kept the same across evaluations.
Agents are evaluated over 50 episodes with the shortcut open and another 50 with it closed, using these preset starting conditions.
Evaluations occur at 36 predetermined checkpoints throughout training, beginning at the adjusted starting points described earlier.
These checkpoints are more concentrated towards the beginning of training where performance changes most rapidly.
At each checkpoint, a copy of the agent's neural network is frozen and tested on the evaluation suite.
This process of evaluating frozen copies of agents allows us to assess not only their navigational abilities, but also to later analyze the activations in their neural networks, which we use to explore learned representations.

\subsection{Representations}
\label{sec:reps}

Representations can be thought of as encodings of environment state information within the activations of an artificial neural network layer \cite{bellemare:2019, lyle:2021}. 
Formally, we may say that the neural network represents a d-dimensional vector feature $x \in \mathbb{R}^d$ if there is a mapping $\phi : \mathbb{R}^n \rightarrow \mathbb{R}^d$ that maps the $n$ activations of a network layer to that feature.
In practice, finding a formal mapping is impractical, so instead we turn to other methods of measuring the information stored in network activity.
We measure how strongly individual nodes encode location information based on how consistently they tend to activate in spatial regions.
Landmark sensitivity is computed based on the differences in node activation with the pink wall landmark being present or absent.
We also examine representations on the population level.
Clustering visualizations are used to qualitatively examine what spatial features are being encoded.
Finally, we measure distances in population activity space to determine how different trajectories are encoded in networks even if they pass through the same spatial locations.
Specific details on how each of these representation measures is calculated is discussed in respective section of the results.

For all discussions of representations, we analyze activities in the recurrent layer of an agent's neural network.
This is the last layer in sequence that is shared between the actor and critic sides of the network.
In principle, representations could be searched for in any layer of the neural network, and other works have performed representation analysis on for example, the penultimate actor layer \cite{bellemare:2019}.
Our focus remains on the recurrent layer due to its role in integrating and storing information, acting as the agent's internal memory, and being shared by the actor and critic branches of the network. These properties make it likely to encode attributes generally useful to both policy and value.

\section{Results}

\subsection{Training conditions affect navigation ability}
\label{sec:performance_curves}

\begin{figure}
\centering
\includegraphics{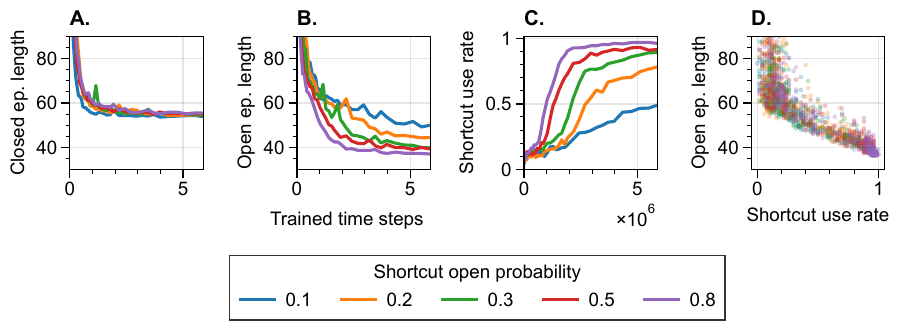}
\caption{
\textbf{(A-C)} Performance curves for agents trained with different probabilities of encountering an open shortcut ($p$). 
\textbf{A.} Mean episode length on closed shortcut episodes. 
\textbf{B.} Mean episode length on open shortcut episodes. 
\textbf{C.} Mean shortcut use rate on open shortcut episodes.
\textbf{D.} Shortcut use rate compared to episode length on open shortcut episodes, with each point representing one agent from one checkpoint.}
\label{fig:p_comb_curves}
\end{figure}

We begin by assessing how efficiently agents are able to learn the shortcut navigation task, with evaluation procedures detailed in the Methods section.
Performance on closed and open shortcut episodes are shown in Fig \ref{fig:p_comb_curves}A and Fig \ref{fig:p_comb_curves}B respectively.
A few key observations can be made from these plots.
First, Fig \ref{fig:p_comb_curves}A demonstrates that regardless of frequency of encountering the open shortcut during training (the value of $p$), agents rapidly achieve optimal performance in closed shortcut trials once initial learning starts. 
Agents trained with $p=0.1$ appear to reach an optimal policy slightly faster than others, due to more frequent exposure to closed shortcut episode.
However, learning speeds in this condition for agents with $p \geq 0.2$ are all roughly equivalent.

Fig \ref{fig:p_comb_curves}B and C show navigation speed and shortcut use rate respectively on open shortcut episodes.
Fig \ref{fig:p_comb_curves}D confirms correlation between the two.
Agents trained in higher $p$ environments gain more experience with open shortcut trials, directly influencing the speed at which they learn to use the shortcut consistently.
Given enough training time, agents can learn to use shortcuts at a rate greater than the probability that they saw shortcuts during training.
For instance, $p=0.1$ agents achieve around $50$\% shortcut usage by the end of training, and agents trained with $p \geq 0.3$ all reach nearly $100$\% shortcut usage.
Given even longer training duration, it is possible that all agents could converge to fully optimal policies.

Performance on open shortcuts is the limiting factor for maximizing rewards, so the remainder of our analysis focuses on shortcut use rate as the primary measure of navigation learning in the simulated RL environment.
In the following sections we analyze learned representations, building up our understanding of what agent networks learn in the navigation environment and how representations influence their performance.

\subsection{Position Representations}
\label{sec:individual_representations}

Drawing inspiration from the hippocampal place cells and grid cells in the brain, which activate selectively based on spatial location \cite{moser:2015}, we initially explore position-based encoding in our artificial agents.
We start by detailing our approach to computing these representations and then develop a measure of spatial representation strength.
Spatial representations develop early in training and subsequently stabilize before navigation strategies fully develop, suggesting that having spatially consistent activations is necessary for basic navigation, but insufficient for advanced strategies.
Then, we use clustering methods on spatial activity heatmaps to qualitatively examine the features used by artificial agents in the shortcut task.
Clustering methods help uncover features in representations that could otherwise be challenging to identify when looking at noisy activation patterns in individual nodes.

\subsubsection{Spatial heatmaps and local variance measure}
\label{sec:spatial_heatmaps}

We first study spatial representations at the individual-node level by generating spatial activation heatmaps.
While an agent is completing the 100 evaluation episodes outlined in Sec \ref{sec:performance_curves}, we collect both the agent's position and the activations of each node in the recurrent neural network layer at each time step. 
We first partition the maze space into a 30 x 30 grid and call the center of each grid cell the grid point $\{g_i\}$. 
Then, to generate the spatial heatmap for a node, we take averages of the node's activities across all collected time steps weighted by proximity between the grid point and the agent at that time. More specifically, 
let $\{x_t\}$ and $\{h_{t,i}\}$ be, respectively, the position of the agent and activity of node $i$ at time $t$.
Each grid point in the heatmap is colored based on a Gaussian spatial average of activity.
Let $w_{t,i} = \exp(-d^2(x_t, g_i) / 2\sigma^2)$ where $d$ is Euclidean distance, then the grid point is colored as a weighted average $\bar{h_{g_i}} = \sum_t h_{t,i}w_{t,i} / \sum_t w_{t,i}$.
In other words, a heatmap illustrates how a node tended to activate when the agent was in different parts of the environment.
Heatmaps are then standardized to have zero mean and unit variance, highlighting deviations from a node's typical outputs rather than absolute activation values.

\begin{figure}
\centering
\includegraphics{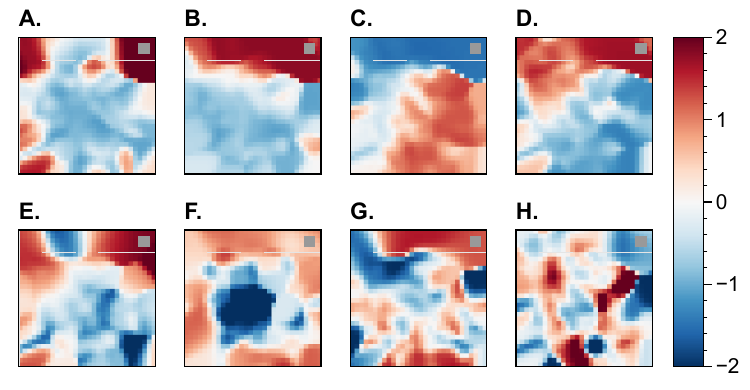}
\caption{Examples of spatial heatmaps generated from agents at different points in training and taken from different $p$ training environments. 
Each heatmap depicts a single node and how it typically activated based on the agent's location.
Heatmaps have zero mean and unit variance, with red and blue indicating where a node had higher and lower than average activation respectively.
Drawings of the goal and walls are also shown in each plot for reference.}
\label{fig:example_spatial_heatmaps}
\end{figure}

Examples of some heatmaps are shown in Fig \ref{fig:example_spatial_heatmaps}.
These are chosen to be representative of common archetypes we observed.
Some nodes show clear sensitivities for certain parts of the maze, such as corners (Fig \ref{fig:example_spatial_heatmaps}A), the corridor (Fig \ref{fig:example_spatial_heatmaps}B, E, G), nearby the corridor entrance (Fig \ref{fig:example_spatial_heatmaps}C, D) or have a center-surround configuration (Fig \ref{fig:example_spatial_heatmaps}F).
They also range in amounts of noisiness, from what we might consider easy to interpret (Fig \ref{fig:example_spatial_heatmaps}A-F) to noisy and difficult to interpret (Fig \ref{fig:example_spatial_heatmaps}G-H).

\begin{figure}
\centering
\includegraphics{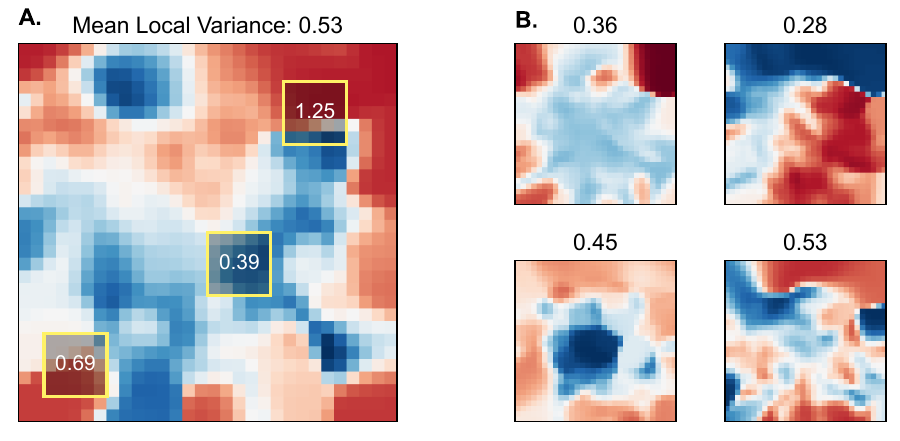}
\caption{
\textbf{A.} Yellow boxes show some example 5x5 windows with the number inside giving the standard deviation of pixels inside.
To get the mean local variance, we take the average of all standard deviations each possible 5x5 window.
\textbf{B.} Some example heatmaps with their corresponding mean local variances shown above.
}
\label{fig:mean_local_var_example}
\end{figure}

To quantify the spatial representation strength of a node, we introduce a measure called the \textit{mean local variance}.
This measures the local consistency of activation patterns within the heatmap.
As illustrated in Fig \ref{fig:mean_local_var_example}A, we calculate standard deviations for each 5x5 pixel window in the heatmap and average these to derive a single score for each node's spatial representation. 
The score effectively captures the intuition that nodes with more consistent and interpretable spatial activations (lower mean local variance) more strongly encode location information
(see Fig \ref{fig:mean_local_var_example}B).
It is likely that a single node is involved in the computation of multiple representations simultaneously, and a noisier heatmap may indicate a node that does not encode spatial information as much as other features.
Note that while this score uses the standard deviation, we call it a local variance, but the two terms are equivalent in interpretation.

\begin{figure}
\centering
\includegraphics{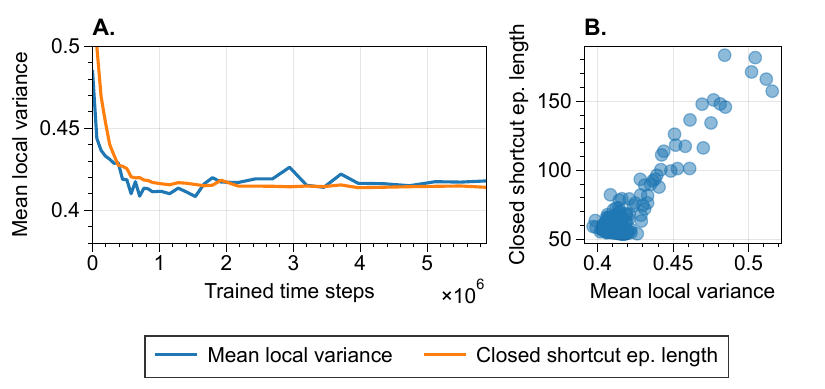}
\caption{
\textbf{A.} Mean local variance shown along with closed shortcut episode lengths for $p=0.1$ agents (agents trained with other $p$ values have similar plots).
Note that episode lengths in this subplot are shown with non-dimensional units and are shown just for comparison of time courses with the local variance measure.
\textbf{B.} Correlation between mean local variance and closed shortcut episode lengths for all $p$ agents.
Each point represents one agent. 
}
\label{fig:local_var_across_training}
\end{figure}

Fig \ref{fig:local_var_across_training}A presents how the mean local variance measure evolves over training for $p=0.1$ agents, alongside the $p=0.1$ closed episode length performance curve from Fig \ref{fig:p_comb_curves}A.
We observe a sharp decline in both measures early in training as agents become proficient at navigation, followed by a plateau, suggesting no further improvement.
A clear correlation between these two is shown in Fig \ref{fig:local_var_across_training}B, and holds across all $p$ agents.
One may think of the closed shortcut performance curves as a proxy for basic navigation skill development since there is only one path to take, and navigation is relatively easy on these episodes.
Spatial representation strength develops simultaneously with closed shortcut performance, suggesting that stable location encodings are important for basic navigation.
In contrast, as we noted in Fig \ref{fig:p_comb_curves}C, performance on open shortcut episodes continues to improve long after these encodings stabilize.
This suggests that spatial information alone is insufficient for optimal policies in the shortcut environment, and as we will see, other representations are also refined with more training.

\subsubsection{Clustering spatial heatmaps}
\label{sec:clustering_spatial_heatmaps}

\begin{figure}
\centering
\includegraphics{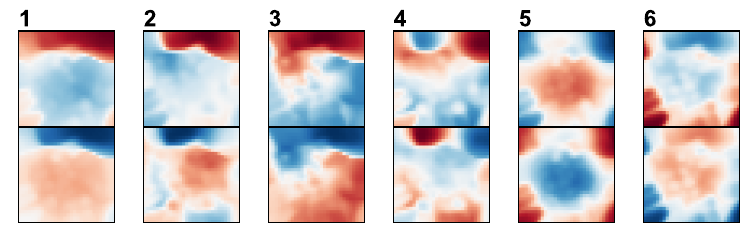}
\caption{
Cluster centers generated from one run of k-means with $k=12$ on spatial activation heatmaps.
Each column shows one ``positive'' cluster center with its closest negative below it.
}
\label{fig:spatial_cluster_centers}
\end{figure}

Beyond quantifying the consistency of single node spatial representations, we are motivated in understanding what qualitative features nodes contribute to encoding.
In order to do so, we demonstrate an effective technique of drawing signal out of the noise by looking at the features that whole populations of nodes encode.
First, we take spatial heatmaps from every agent across all $p$ values and 36 checkpoints, then apply k-means clustering with $k=12$. 
Despite slight variations when running multiple iterations of k-means, the clustering algorithm consistently settles on similar features.
Notably, these clusters partition such that, when multiplying a cluster center by $-1$ and inverting its sign, it approximates another cluster.
This phenomenon is demonstrated in Fig \ref{fig:spatial_cluster_centers} which depicts the cluster centers from one run of k-means, each of which can be visualized by a heatmap (the average of all heatmaps belonging to the cluster).

\begin{figure}
\centering
\includegraphics{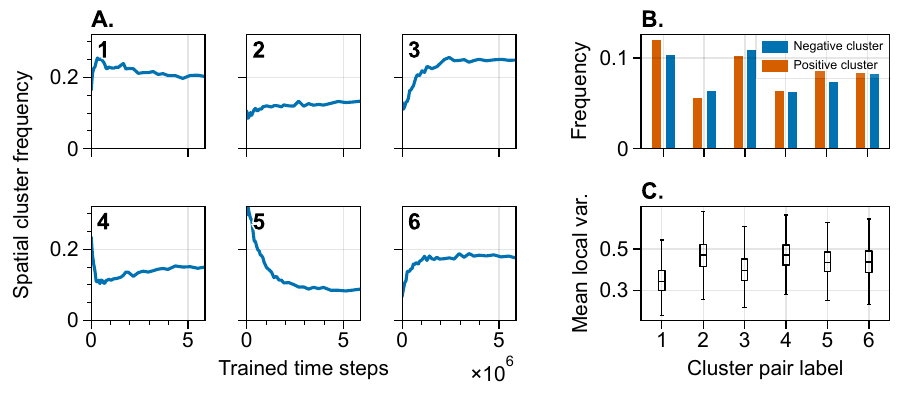}
\caption{
\textbf{A.} Spatial heatmap cluster labeling frequency over training.
\textbf{B.} Comparison of frequencies of positive and negative cluster occurrences (where positive and negative are arbitrarily labeled as in Fig \ref{fig:spatial_cluster_centers})
\textbf{C.} Mean local variance measure for spatial heatmaps with a given cluster label.
Boxplots are centered at the median, with the body going from the first quartile (Q1) to the third quartile (Q3).
Whiskers are drawn at 1.5 inter-quartile ranges below and above Q1 and Q3 respectively.
}
\label{fig:spatial_kmeans_comb}
\end{figure}

Note that the definitions of positive and negative clusters are arbitrary.
For a given positive negative cluster pairing, the frequency of nodes being labeled as positive or negative are roughly equal, as depicted in Fig \ref{fig:spatial_kmeans_comb}B. 
This relationship reflects the neural network's ability to transmit the same information using activations whether they are above or below a node's average.

Some features are clearly interpretable in the cluster centers of Fig \ref{fig:spatial_cluster_centers}. 
For example, clusters 1 and 2 are sensitive to the corridor of the maze.
Cluster 3 extends this sensitivity to the long-path entrance to the corridor.
Clusters 4 and 5 are sensitive to the top-right corner where the goal is located, suggesting of place-field-like encoding.
Finally, cluster 6 looks like an edge-detection feature.

Fig \ref{fig:spatial_kmeans_comb}A tracks these cluster frequencies across training, revealing stabilization early on for some clusters.
Cluster 5 frequency (the cluster with clearest corner representation) starts the highest and drops off.
Nodes in this cluster likely act as a place field for the goal, which may be important when the agent is first learning what the goal is.
However, as the agent learns the routes to the goal, the utility of knowing the precise goal location decreases.
Conversely, the frequency of nodes belonging to Cluster 3 that are sensitive to the long-path entrance see an increase over training, underscoring their navigational value.

The mean local variance scores of nodes in different clusters are mostly comparable, as shown in Fig \ref{fig:spatial_kmeans_comb}C.
We emphasize that the boxplots show large ranges of mean local variances of constituent nodes (See Fig \ref{fig:mean_local_var_example} for some pictures of what heatmaps with different local variance scores look like).
Every cluster contains nodes ranging from locally consistent to noisy.
Despite this large variability of spatial consistency in each cluster, the features illuminated by the clustering method are consistent across k-means runs. 
One might imagine that while individual nodes may be spatially noisy, by combining inputs from multiple nodes these features may be effectively transmitted through the network.

\begin{figure}
\centering
\includegraphics{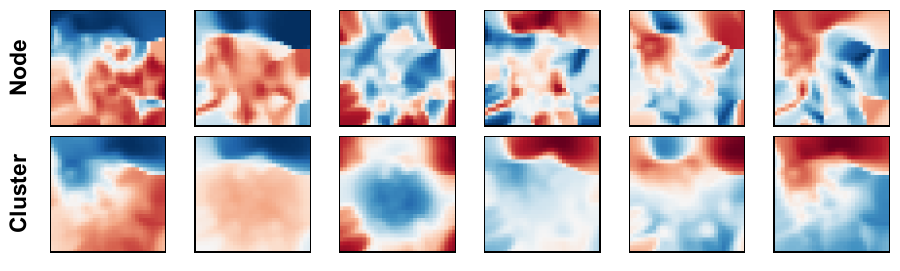}
\caption{Examples of some spatial heatmaps with the corresponding cluster centers that they were assigned to.
Each column corresponds to one node, with the node's activity heatmap shown on the top, and the cluster center on the bottom.
Nodes were selected from different agents.}
\label{fig:spatial_map_cluster_examples}
\end{figure}

To further demonstrate this idea, in Fig \ref{fig:spatial_map_cluster_examples} we show a few spatial heatmaps alongside the clusters they are classified into (visualized by their centers).
By comparing the two, we can qualitatively distinguish the most notable features of a heatmap image, suggesting what information a node may be involved in providing to downstream computations in the network.
This clustering approach enhances our understanding of spatial encodings by artificial agents, highlighting how their strategies evolve from using place-field-like behavior and tracking goal location, to using the corridor entrance as a key navigation feature.
It also provides context to interpret the activations of nodes that might otherwise be noisy and difficult to interpret.
Clustering methods of this sort might have broader applicability to interpreting neuron recordings in biological studies, where data is often noisy, and in general serves as a useful explorative tool to determine what representations are used by agents.
Later in Sec \ref{sec:population_representations}, we give another example of the effectiveness of analyzing populations of nodes in uncovering representations.

\subsection{Landmark Sensitivity Representation}
\label{sec:landmark_sensitivity}

The next representation we consider is landmark sensitivity, specifically measuring how responsive a node is to changes in an available visual landmark.
Recall that the shortcut wall is colored pink unlike other white walls in the environment, and it acts as a salient cue when closed off.
This leads to an interesting feature of our task environment where cue availability and shortcut opening are coupled together.
We hypothesized that having nodes with a robust representation of this landmark should correlate to an agent's ability to recognize and utilize the open or closed state of the shortcut, and hence correlate with increased performance.
This concept is inspired by studies in humans, where the use of distal and proximal cues is important in navigation skills \cite{chen:2009, siegel:1975}.
As we will discuss in this section, landmark encodings are improved by use as part of navigation strategies, and that this frequent exposure alone does lead to development of a robust landmark representation.

\begin{figure}
\centering
\includegraphics{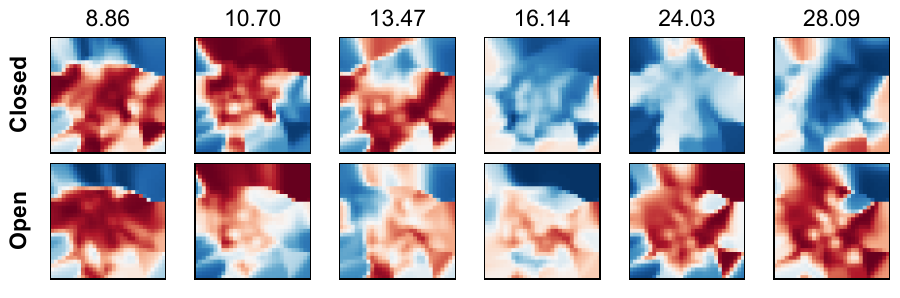}
\caption{Examples of landmark sensitivity scores for a few nodes drawn from one $p=0.5$ agent late in training.
Each column of 2 heatmaps are the spatial activity heatmaps for a single node from 50 closed shortcut episodes (top) and 50 open shortcut episodes (bottom).
The number above the heatmap is the landmark sensitivity score for the node, which is calculated as the absolute difference of the two heatmaps.
}
\label{fig:landmark_sensitivity_examples}
\end{figure}

To quantify a node's sensitivity to the pink wall landmark, we compare its spatial activity heatmap from 50 closed shortcut episodes with the heatmap generated from 50 open episodes.
A large difference between these two indicates high landmark sensitivity.
This measure could alternatively be assessed by comparing average node activations between open and closed episodes, but we use spatial heatmaps which offer a clear visual distinction in node responses under open and closed shortcut states, as seen in examples presented in Fig \ref{fig:landmark_sensitivity_examples}.
We note that with a landmark score above around $20$, heatmaps are clearly distinct, such as the two right-most nodes shown in the examples.

\begin{figure}
\centering
\includegraphics{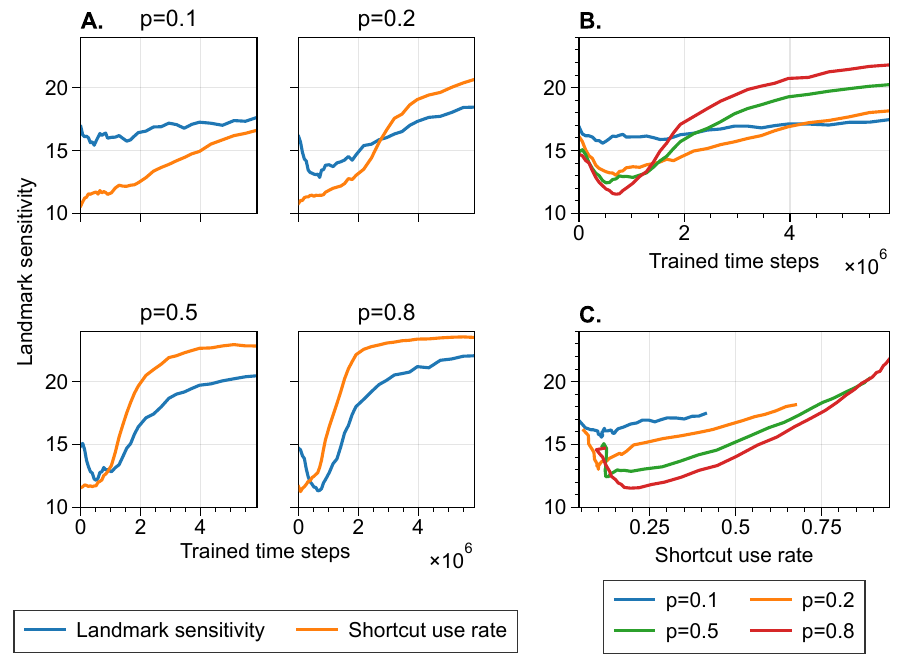}
\caption{
\textbf{A.} Comparison of shortcut use rate through training with cue sensitivity scores for individual $p$ value agents.
\textbf{B, and C} both show combined results on the same axis for multiple $p$ values for comparison.
\textbf{B.} Cue sensitivity across training time.
\textbf{C.} Visualization of how cue sensitivity changes along with shortcut use rate. 
}
\label{fig:landmark_sensitivity_combined_results}
\end{figure}

Two key observations arise from this landmark sensitivity representation score.
First, as shown in the subplots of Fig \ref{fig:landmark_sensitivity_combined_results}A and in the combined plot Fig \ref{fig:landmark_sensitivity_combined_results}B, there is an initial decrease in landmark sensitivity during the early stages in training, particularly for agents trained with $p > 0.1$.
This period coincides with the same early period of training where spatial representations develop (as seen from the mean local variance in Sec \ref{sec:spatial_heatmaps}).
We hypothesize that the natural tendency of neural networks in this task is to first over-emphasize developing location encodings at the cost of other information.
Frequent exposure to the landmark however, is preventative for this loss, indicated by a smaller dip in landmark sensitivity the lower $p$ is and the more often and agent sees the landmark.
However, we also note that as training progresses, higher $p$ agents that progress in navigation skill more quickly eventually overtake lower $p$ agents in landmark representation scores.


Our second observation focuses on the relationship between shortcut use rates and landmark sensitivity scores, shown in Fig \ref{fig:landmark_sensitivity_combined_results}C.
Towards the end of training, higher p agents tend to converge toward optimal performance (100\% shortcut use rate) and consistently develop average landmark sensitivity scores above 20 (the plotted lines converging at the top right of the plot in Fig \ref{fig:landmark_sensitivity_combined_results}C) regardless of $p$ value.
With even more training, we expect agents with lower $p$ values of $0.1$ and $0.2$ might also develop towards this optimum.
This convergence indicates that the development of landmark sensitivity is tightly coupled with shortcut taking strategies in our environment.

To conclude, we first observed a dip in landmark representation strength early in training, but increased exposure to the landmark (lower $p$) mitigated the severity of this dip.
We also noted that the landmark sensitivity is tightly coupled with shortcut use frequency, and higher $p$ agents that learn to use the shortcut more effectively surpass lower $p$ agents in landmark sensitivity scores after training.
These insights suggest that while exposure to landmarks aids in developing representations, effectively integrating them into navigation strategies is more influential to building robust representations than consistent availability of the cue is.

\subsection{Population level representations and agent ``planning''}
\label{sec:population_representations}

In this section, we expand our focus from individual node representations to those spanning an entire network layer. 
We use the same recurrent layer activations as before but analyze the combined activations of all nodes in the layer.
Our preliminary exploration involved applying the Uniform Manifold Approximation and Projection (UMAP) nonlinear dimensionality reduction technique \cite{mcinnes:2018}, reducing the set of T 64-dimensional vectors (where T is the total number of time steps of data collected in all 100 evaluation episodes, and $64$ is the number of nodes the layer) to T 2-dimensional vectors that can be visualized.
An example is given in Fig \ref{fig:trajectory_kmeans_example}A. Different points here represent network activities in the recurrent layer at different time steps. UMAP attempts to preserve local distances, so the relative proximity of points indicates similarity of population activities at those time instances. 

To explore the structure of the population-level activations, we initially applied k-means clustering to the complete set of T 2-dimensional activations (different clusters represented by different colors in Fig \ref{fig:trajectory_kmeans_example}A).
We then mapped out the positions of the agent corresponding to each clustered point.
Fig \ref{fig:trajectory_kmeans_example} gives a visual demonstration of the process.
This k-means clustering revealed that in well-trained agents, clusters frequently emerged that corresponded distinctly to different segments of the agent's trajectory.
The clearest separation of clusters in UMAP space distinguished times when the agent was above the corridor from points below, with clusters rarely containing both types of points.
Additionally, clusters often formed such that points from trajectories where the agent took the long path and trajectories where the agent took the shortcut were placed in separate clusters (see example Fig \ref{fig:trajectory_kmeans_example}). Note that cluster membership was assigned based only in the similarity of population activity (after UMAP projection). Corresponding similarity of spatial locarions of the agent was found as a consequence.

These patterns formed the basis of a population-level trajectory representation which we explore further in the following discussion.
This approach enables a broader understanding of how collective activations of neural network nodes correlates with an agent's navigational decisions and strategic ``planning'' during navigation tasks.

\subsubsection{Quantifying trajectory set separation scores}

\begin{figure}
\centering
\includegraphics{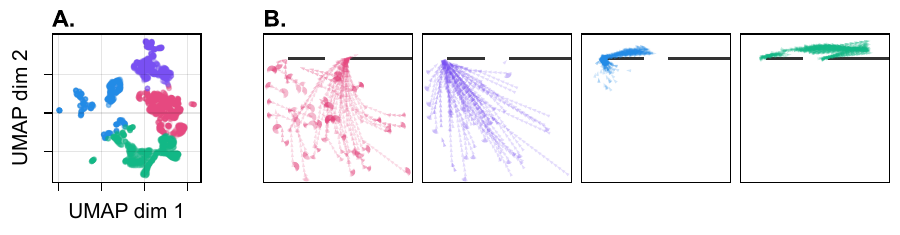}
\caption{Example of k-means clustering data points from a $p=0.1$ agent late in training.
\textbf{A.} Activations from the recurrent network layer reduced to 2 dimensions via UMAP and clustered according to k-means with $k=4$.
\textbf{B.} Each subplot shows where the agent was at time points corresponding to the clustered activation points with larger markers demarcating the initial positions of trajectories.
}
\label{fig:trajectory_kmeans_example}
\end{figure}

\begin{figure}
\centering
\includegraphics{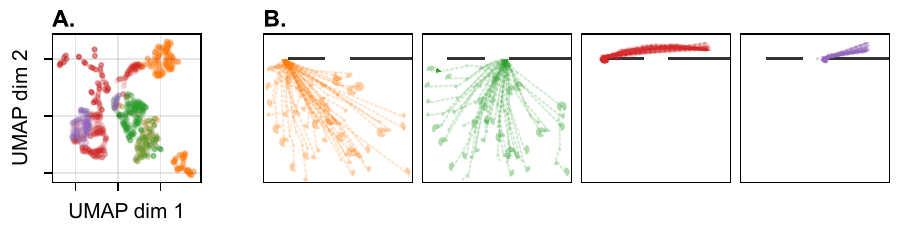}
\caption{
An example of population activity and position data from a $p=0.8$ agent late in training according to predefined trajectory sets.
\textbf{A.} Activations collected over 50 open and 50 closed episodes of an agent performing its policy, reduced to 2 dimensions via UMAP for visualization.
(Note that although hard to see, there is a region of overlap between green and orange points slightly towards the bottom-right of the plot.)
\textbf{B.} Each subplot shows where the agent was at time points belonging to predefined sets.
From left to right: pre-entrance, pre-shortcut, post-entrance, post-shortcut.
}
\label{fig:trajectory_cluster_example}
\end{figure}

To look further into the relationship between distinct population activation vectors and spatial locations, including those in not-yet-well-trained agents, we introduce a population-level representation score to quantify the idea of network activity separating in activation space based on the parts of a trajectory an agent was on.
First, we separate time steps into 4 sets based on the agent's trajectory Fig \ref{fig:trajectory_cluster_example}, following the intuition gained from the previous clustering exploration. 
Positions of the agent for each set are displayed in Fig \ref{fig:trajectory_cluster_example}B.
From left to right, the first two sets contain points where the agent was below the corridor, and we call them ``pre-entrance'' (colored orange) if the agent took the longer corridor entrance path on that episode, or ``pre-shortcut'' (colored green) if the agent took the shortcut.
The next two sets contain points where the agent was in the corridor, and similarly are termed ``post-entrance'' and ``post-shortcut'', depending on which path the agent took on that episode.
The set definitions are independent of whether the shortcut was actually open on a given episode.
These sets aim to capture the ``intent'' or ``plan'' of an agent, rather than being based only on its spatial position.
Positions in the ``pre-entrance'' and ``pre-shortcut'' sets, for example, have similar spatial points, but their activations in UMAP space can often be separated as seen in the green and orange points in Fig \ref{fig:trajectory_cluster_example}A.
Note that Fig \ref{fig:trajectory_cluster_example} shows the same set of points as Fig \ref{fig:trajectory_kmeans_example} but with grouping based on predetermined trajectory clusters instead of unsupervised clusters.

To measure the separation of these sets in activation space, we employ the Wasserstein distance, a metric that calculates the minimum sum of Euclidean distances between paired points in two sets.
Let $\{u_i\}$ be the points belonging to a set $U$, and likewise for $\{v_j\}$ and $V$, where $u_i$ and $v_j$ are 64-dimensional activation vectors.
The Wasserstein distance between $U$ and $V$ is defined as the sum of distances between pairs $(u_i, v_j)$ such that the distance $\tilde{W} = \sum_{i, j} |u_i - v_j|$ is minimized.
Each point is paired at most once, and all points from the smaller set are included in the calculation (some from the larger set may be left out).
We normalize the Wasserstein distance by $W = \tilde{W}/(N \cdot D)$ where $N$ is the number of pairings, and $D = \max |x_i - y_i|$ is the maximum Euclidean distance for any two points in the entire activation space.

We find that the most interesting set pair is the pre-entrance and pre-shortcut pair (the orange and green points in Fig \ref{fig:trajectory_cluster_example}) and is the \textit{set separation score} that will be the main focus of our analysis.
The separation score of this pair increases the most across training and has the best correlation with shortcut usage.
In the example shown in Fig \ref{fig:trajectory_cluster_example}A, there are some distinct regions that only contain orange points in the top-right and bottom-right of the plot.
There is also a region of overlap between the two clusters of interest slightly towards the bottom-right of the image.
In this example, the sets separate quite well despite containing many points with similar positions.

\begin{figure}
\centering
\includegraphics{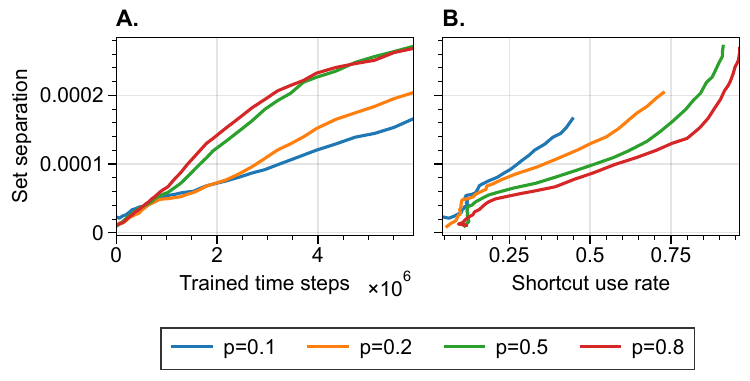}
\caption{
Wasserstein distances between pre-entrance and pre-shortcut network activation clusters.
\textbf{A.} Mean Wasserstein distances across training.
\textbf{B.} Visualization of how Wasserstein distance changes with respect to performance
}
\label{fig:trajectory_cluster_separation}
\end{figure}

Fig \ref{fig:trajectory_cluster_separation}A tracks the development of this set separation score across training.
There is a clear positive association between the set score and shortcut usage, as seen in Fig \ref{fig:trajectory_cluster_separation}B, and unlike landmark sensitivity, this representation measure increases in a monotonic fashion.
The separation of pre-shortcut and pre-entrance sets indicate that the population activity contains information about the intended route of the agent, and not merely the agent's current position.
Furthermore, the correlation of higher separation scores with improved performance indicates that this population-level representation which separates trajectories or strategies is useful in advanced navigation.
We will inspect this idea more carefully in the next section.

\subsubsection{Prescribed trajectories}
\label{sec:prescribed_trajectories}

\begin{figure}
\centering
\includegraphics{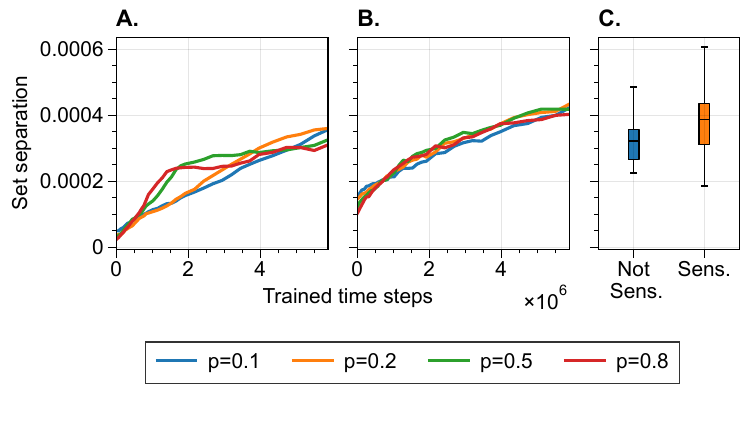}
\caption{
Wasserstein distances between activations from pre-entrance and pre-shortcut clusters.
\textbf{A.} Mean Wasserstein distances only considering episodes where the shortcut was open.
\textbf{B.} Mean Wasserstein distances using prescribed trajectories, open shortcut only.
\textbf{C.} Boxplots showing Wasserstein set separation scores (policy trajectories, open episodes only) late in training while only considering cue sensitive nodes or non-cue sensitive nodes in sets.
}
\label{fig:trajectory_open_prescribe_combo}
\end{figure}

We interpreted the pre-shortcut and pre-entrance separation score increasing as an indication of a neural network's growing effectiveness in encoding trajectory or planning information. 
However, an alternate interpretation could be that this increase is not purely due to enhanced encoding abilities, but rather due to changes in policy during training that lead to modified visual inputs.
For instance, an agent early in training that only takes shortcuts sometimes will end up with open shortcut episodes in both pre-entrance and pre-shortcut sets.
On the other hand, an optimized, well-trained agent might have only open episodes in the pre-shortcut set and closed episodes in the pre-entrance set.
The well-trained agent then would have more differing visual stimuli between the two sets, due to the landmark only occurring when the shortcut is closed.
One might conclude that an increase in set separation after training is observed due to policy optimization that causes visual stimuli received in the two sets to differ.

Countering this interpretation, we observe that the set separation score continues to rise for $p=0.8$ agents even after performance has optimized and plateaued (at around $2e6$ timesteps, $p=0.8$ agents use shortcuts about $90\%$ of the time, from Fig \ref{fig:p_comb_curves}C).
This suggests that the population representation continues to strengthen despite stable visual stimuli.
This is further evidenced in Fig \ref{fig:trajectory_open_prescribe_combo}A, where we analyze separation scores exclusively using data from open shortcut episodes.
The positive correlation between score and skill is still preserved, lending further evidence that agents are learning to encode trajectory information, and not merely reacting to differential visual stimuli.

To conclusively demonstrate that increased set separation is not only a reflection of changes to behavior, we employ a novel network activity collection strategy of prescribed trajectories.
In this method, we copy the action history of one agent that had approximately $60\%$ shortcut use rate, and then collect activations for all agents under those same actions.
This allows us to observe how agents respond to a fixed set of behaviors and visual stimuli.
Fig \ref{fig:trajectory_open_prescribe_combo}B illustrates that, even with fixed trajectories, agents progressively learn to distinguish between paths leading to the entrance versus the shortcut.

Interestingly, in both alternate measures of set separation shown in Fig \ref{fig:trajectory_open_prescribe_combo}A and \ref{fig:trajectory_open_prescribe_combo}B, the difference between different $p$ agents is diminished or eliminated.
This suggests that a more optimized policy does contribute to increased set separation score, along with better representation encoding.
The key takeaway from this section is that agents learn representations at the population level that are indicative of what trajectory they are headed towards, rather than just their current location.
Although closely associated with shortcut usage, these representations also evolve independently of performance, continuing to mature even when policies are nearly optimal.
Future research could explore the underlying mechanisms driving this continued representation development, potentially looking into what drives network gradients once earned reward signals have stabilized.

\subsubsection{A role of landmark sensitive nodes in trajectory representations}
\label{sec:landmark_in_trajectories}

In Sec \ref{sec:landmark_sensitivity}, we categorized nodes with a landmark representation score greater than $20$ as being sensitive to the shortcut wall landmark.
Fig \ref{fig:trajectory_open_prescribe_combo}C demonstrates that these landmark-sensitive nodes contribute more significantly to the separation of pre-shortcut and pre-entrance activations compared to non-sensitive nodes.
This analysis only includes open shortcut episodes to minimize the impact of varying visual stimuli on the result.

This finding suggests that even in the absence of the landmark, landmark-sensitive nodes play a role in shaping population-level representations.
This could indicate their general importance in advanced navigation or trajectory planning.
Alternatively, the landmark sensitivity measure might effectively identify nodes that consistently provide valuable navigational information to the rest of the network.
It is not uncommon to observe some activations in a neural network layer to become less informative over RL training \cite{lyle:2023, sokar:2023}, and representation scores may be a way to filter out ``useful'' nodes.
These observations may warrant further future study, potentially by exploring additional quantifiable representations on individual nodes to see if nodes scoring highly on those metrics also play pivotal roles in population codes.

To summarize Sec \ref{sec:population_representations}, our detailed analysis of trajectory representations implies that agent networks learn to encode information about intended trajectories.
Additionally, the correlation between this score with performance is due to more than just differences of visual stimuli between sets.
Even though many individual nodes have a strong spatial component, we can use population activity to derive predictive signals for the agent's behavior/policy.
This perspective of looking beyond individual spatially-tuned nodes to sets could be useful beyond artificial agents, and may have value in understanding collections of spiking neural data, in particular to look for encodings that may be hard to find in individual neurons.

\subsection{Comparisons with human navigators}
\label{sec:human_predictions}

\begin{figure}
\centering
\includegraphics{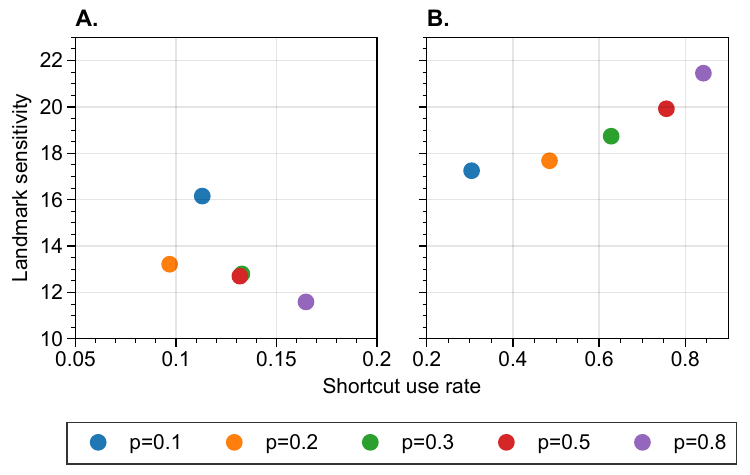}
\caption{Comparison of landmark sensitivity scores with shortcut use rate at \textbf{A.} early in training, near the point of lowest cue sensitivity; \textbf{B.} near the end of training}
\label{fig:landmark_sensitivity_set_points}
\end{figure}

In this final section, we attempt to derive insights into human navigators based on findings from our analysis of artificial agents.
Our research was inspired by a study examining how the environment individuals live in affect their navigation skills and strategies \cite{barhorst:2021}.
Barhorst-Cates et al. assessed individuals from Salt Lake City and Padua on a number of different tasks, including the Dual Solutions Paradigm (DSP), designed to evaluate shortcut usage.
Individuals from Padua were found to have a greater ability to use proximal cues during navigation than those from Salt Lake City as well as use shortcuts more often in the DSP.

While acknowledging that our toy navigation environment is abstracted from the real world, we compare our $p=0.1$ agents to individuals from Salt Lake City, where grid-like street layouts offer few shortcuts and mountains offer persistent global navigation landmarks.
On the other hand, Padua's irregular and organic street layouts are rich in shortcut opportunities and is likened to the $p=0.8$ variation.
This comparison matches learning curves shown earlier in Fig \ref{fig:p_comb_curves} where the higher the $p$ an agent was trained in, the faster its performance developed.
Barhorst-Cates et al. similarly found that individuals from Padua to have generally better navigation skills than those from Salt Lake City.
With this framework, we make two predictions for human navigators and the environments that they live in.

\subsubsection{A small increase in shortcut experience induces a large learning benefit}

Referring back to Fig \ref{fig:p_comb_curves}C, we observed a large gap in learning efficiency between agents from $p=0.1$ and $p=0.8$ environments.
One might assume $p=0.2$ agents would resemble  those from $p=0.1$, with closed shortcut episodes dominating their experience.
Yet, their learning dynamics and representation development, particularly the initial dip in landmark sensitivity, are more qualitatively similar to $p=0.5$ or $p=0.8$ agents.
Additionally, $p=0.3$ agents reach nearly the same asymptotic performance as agents in $p=0.8$, far beyond those from $p=0.1$. 
This suggests that even small adjustments in training conditions can significantly influence learning outcomes.

We propose a modification to the DSP to test this hypothesis.
Typically, DSP participants are first instructed to walk around the perimeter of a maze multiple times while noting the objects that they come across.
Afterwards, they are tasked with finding their way to these objects, and the frequency that they use shortcuts in the maze (cutting through the maze, as opposed to following the path they were initially guided on) is recorded.

During the initial phase of guided exploration, a modified DSP task could incorporate some number of example paths where participants are led through the center of the maze.
We hypothesize that these demonstrations should induce a nonlinear improvement in the amount of shortcuts used by participants as a function of the number $N$ of shortcut demonstrations given.
In other words, as $N$ increases from $0$, we predict that initially a substantial increase in shortcut taking will be observed.
Eventually the marginal benefit of increasing $N$ further should taper off, with the largest gains in shortcut usage coming from the first few added demonstrations.

\subsubsection{Dependence of representation and navigation skill development on environment}

Landmark knowledge is usually regarded as beneficial for navigation in humans \cite{chen:2009, siegel:1975}.
In our artificial agents, we showed that this was certainly the case, where optimal navigation policies and robust landmark sensitive representations developed concurrently (see Fig \ref{fig:landmark_sensitivity_combined_results}C), particularly in agents trained with higher $p$ values.
Similarly, individuals from Padua were found to have better allocentric navigation skills and more ability to use proximal cues in navigation than those from Salt Lake City \cite{barhorst:2021}.
However, the amount of experience an individual had was not a main focus in that study.

In our deep RL agents, we found that amount of training modulated the relationship between shortcut use and landmark sensitivity for different $p$ values.
This result is highlighted in Fig \ref{fig:landmark_sensitivity_set_points}, which depicts the relationship early and late in training.
Early on, lower $p$ agents showed better representation scores, due to the previously discussed dip in score that is mitigated by exposure to the landmark.
However, as training progresses, the ordering reverses with higher $p$ agents achieving higher representation scores.
This suggests a prediction in human navigators, that amount of experience may modulate the relationship between growing up environment and landmark knowledge.

Specifically, to test this hypothesis one could conduct a factorial analysis of previously collected data, where individuals from different environments are subdivided based on navigational skill.
Navigation ability would act as a surrogate measure for amount of experience.
We would predict that similar to the artificial agents trained on the simulated shortcut task, better landmark knowledge would be found in individuals from an environment with stable global cues but less experience.
When looking at the group with more experience however, we would expect to find better landmark knowledge in those from the more challenging navigation environment.



\section{Discussion}

Our work is motivated by research into individual differences in navigation skill in humans.
We used deep RL agents in a toy navigation task as a model for behavioral studies of human navigators while also taking inspiration from navigation research to develop RL training environments and representation analysis techniques.

Our study of learned representations in network nodes' activity began with inspiration from place cells \cite{moser:2015}, mapping out locations that individual nodes were sensitive to.
We also found nodes that were sensitive to visual stimuli provided by the pink wall cue in the simulated maze, noting that the strength of this representation was dependent on both frequency of seeing the cue, and also skill in using it as part of navigation.
While the artificial agents we developed are detached from the complexities of the real world, they provide a useful model for learning, especially because of the granular access we have to the activity patterns driving their behavior.
Our findings led us to predictions about how human navigators may develop internal representations and components of navigation knowledge.
Specifically, we suggest that skill or amount of experience may be a key modulating factor in what is learned from different environments, and that general trends (such as one population having better landmark knowledge than another) may obfuscate more nuanced differences in individuals.
A large portion of our findings come from finding ways to quantify the strength of representations, and then looking at how these scores change over training.
In general, this type of strategy may be useful to explore the depth of how representations develop over time in both neural networks and brains alike.

In this work, we also develop techniques to analyze encodings in populations of node rather than individual neurons.
Clustering in particular was used as a tool for preliminary exploration and provides qualitative suggestions of features that are important to agents.
In Sec \ref{sec:clustering_spatial_heatmaps}, k-means revealed common spatial features that developed in navigation learning.
When matching a heatmap to its nearest cluster, it becomes much clearer what feature that node might help encode.
In Sec \ref{sec:population_representations}, applying k-means directly to UMAP-reduced activity data revealed the existence of a trajectory-based representation different from the position ones we had originally set out to explore.
The key takeaway is that looking at the population of activities or neural firing rates may be a useful exploratory tool to better understand what a network or brain is encoding.
In the context of the whole population of activity, noisy individual node activations reveal themselves to be part of a larger picture of useful representations, and this perspective may be broadly valuable in analyzing neural recordings.

Our primary approach to collecting network activity data was to record activations while the agent performed its usual policy on a set of evaluation trials.
This comes with an important caveat that changes to behavior could lead to shifts in observations input to the network, leading us to mistake behavioral changes to changes in representations.
As demonstrated in Sec \ref{sec:prescribed_trajectories}, with artificial agents we have the option of feeding fixed, repeatable stimuli to the agent and seeing how the network responses change through training.
However, we have noticed that controlling stimuli can diminish the strength of representations, and could lead to some encodings being missed entirely.
In artificial agents, since representations are optimized as part of an agent's policy, they should be considered valid within that context.
Place cells, which serve as a main motivation, have also been found to be dependent on environment and task context \cite{kobayashi:1997, mcnaughton:2006}.
Thus, using prescribed trajectories is a valuable technique to control for the influence of policy changes in representation development, but representations should also be analyzed with the policies they are developed.

There are many directions that could further this research. 
As discussed in Sec \ref{sec:prescribed_trajectories}, the trajectory encoding continues to strengthen even after an agent's policy has stabilized.
It would be interesting to unpack why these changes happen, possibly by examining gradients driven by the RL algorithm.
Without behavioral changes, it is somewhat surprising that network representations would continue to develop.
On the idea of whether certain nodes with strong individual representations are more important to behavior discussed in Sec \ref{sec:landmark_in_trajectories}, we could test the influence of these nodes on policy by performing ablation studies of them.
Finally, to further validate the idea of agents in high $p$ environments developing stronger cue sensitivity with training, one could design transfer experiments where trained agents are moved to another navigation context that requires landmark usage.

{\bf Acknowledgements.} 

This research was partially funded by the National Science Foundation. We thank Dr.\ Sarah Creem-Regehr, Dr.\ Elliot Smith, Dr.\ Frederick Adler, and Dr.\ Akil Narayan for helpful discussions of this work. The support and resources from the Center for High Performance Computing at the University of Utah are also gratefully acknowledged.

\bibliographystyle{plain}
\bibliography{refs}

\end{document}